# COMPARISON OF DIFFERENT T-NORM OPERATORS IN CLASSIFICATION PROBLEMS

Fahimeh Farahbod[1] and Mahdi Eftekhari[2]

[1]Department of Computer Engineering, Shahid Bahonar University, Kerman, Iran
*f.farahbod@eng.uk.ac.ir*
[2] Department of Computer Engineering, Shahid Bahonar University, Kerman, Iran
*m.eftekhari@uk.ac.ir*

## ABSTRACT

*Fuzzy rule based classification systems are one of the most popular fuzzy modeling systems used in pattern classification problems. This paper investigates the effect of applying nine different T-norms in fuzzy rule based classification systems. In the recent researches, fuzzy versions of confidence and support merits from the field of data mining have been widely used for both rules selecting and weighting in the construction of fuzzy rule based classification systems. For calculating these merits the product has been usually used as a T-norm. In this paper different T-norms have been used for calculating the confidence and support measures. Therefore, the calculations in rule selection and rule weighting steps (in the process of constructing the fuzzy rule based classification systems) are modified by employing these T-norms. Consequently, these changes in calculation results in altering the overall accuracy of rule based classification systems. Experimental results obtained on some well-known data sets show that the best performance is produced by employing the Aczel-Alsina operator in terms of the classification accuracy, the second best operator is Dubois-Prade and the third best operator is Dombi. In experiments, we have used 12 data sets with numerical attributes from the University of California, Irvine machine learning repository (UCI).*

## 1. INTRODUCTION

In recent years, fuzzy models have been used widely because they are able to work with imprecise data and acquired knowledge with these models is more interpretable than the black-box models. Fuzzy models are able to handle the complex nonlinear problems. The fuzzy modelling process has generally intended to deal with an important trade-off between the accuracy and the interpretability of the model. Recently, tendency to look for a good balance between the accuracy and the interpretability has increased. Fuzzy Rule-Based Classification System (FRBCS) is a special case of fuzzy modelling. FRBCS focuses on finding a compact set of fuzzy if-then classification rules to model the input-output behaviour of the system. The input of the FRBCS is a number of pre-labelled classification examples, and the output of this system is a crisp and discrete value. One important advantage of a FRBCS is its interpretability.

FRBCS is composed of three main components: database, rule-base and reasoning method. The database contains the fuzzy set definitions related to the linguistic terms used in the fuzzy rules. The rule base consists of a set of fuzzy if-then rules in the form of "if a set of conditions are satisfied, then a set of consequences can be inferred". Reasoning method uses information from database and rule-base to determine a class label for patterns and to classify them.

In this work, we compare the effect of the 9 most widely used *T*-norm operators on the accuracy of FRBCS. We use a simple and efficient heuristic method for constructing FRBCS. Let us assume that our pattern classification problem is a *n*-dimensional problem with *C* classes and *m* training patterns, $X_p = [x_{p1}, x_{p2}... x_{pn}]$, $p = 1, 2 ...m$. Usually, each attribute of the given training patterns is normalized into a unit interval [0, 1] by using a linear transformation that preserves the distribution of training patterns. We used 14 fuzzy sets showed in Fig. 1 to partition the domain interval of each input attribute. Triangular shaped fuzzy sets are used, because they are simple and more human understandable [1]. Each fuzzy rule should use one of these fuzzy sets to specify the value of each attribute.

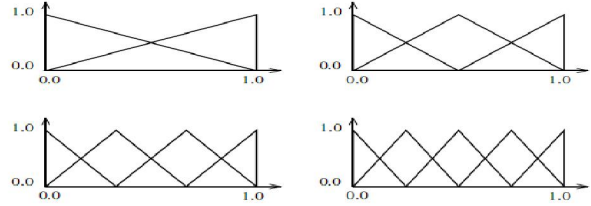

Figure 1. Different partitioning of each attribute axis [1].

Let us assume $X_1 = [x_{11}, x_{12} ... x_{1n}]$ is the input attribute vector, $R_q$ is the label of the *q-th* fuzzy if-then rule, $A_{q1}, A_{q2} ... A_{qn}$ are antecedent fuzzy sets on the unit interval [0, 1], $C_q$ is the consequent class, $CF_q$ is the certainty grade of rule $R_q$ (i.e. weight of rule). We have used fuzzy rules of following type:

*if* $x_{11}$ *is* $A_{q1}$ *and ...and* $x_{1n}$ *is* $A_{qn}$ *then Class* $C_q$ *with* $CF_q$.

In the field of data mining two measures confidence and support are frequently used for evaluating fuzzy rules. In order to classify an input pattern $X_p = [x_{p1}, x_{p2}... x_{pn}]$, the compatibility degree of the pattern with each rule is calculated. For calculating the compatibility degree of the pattern with each rule and calculating confidence and support of each fuzzy rule we have used 9 different *T*-norm operators. In case of using product as *T*-norm operator to model the "and" connectives in the rule antecedent, the compatibility degree of pattern $X_p$ with the rule $R_q$, confidence and support of the rule can be calculated by (1).

$$\mu_q(X_p) = \prod_{i=1}^{n} \mu_{qi}(X_{pi}), \qquad (1)$$

Where $\mu_{qi}(.)$ is the membership function of the antecedent fuzzy set $A_{qi}$. Confidence (denoted by Conf) and support (denoted by Supp) of a fuzzy rule are defined respectively by (2) and (3).

$$Conf(A_q \Rightarrow Class\ C_q) = \frac{\sum_{X_p \in Class\ C_q} \mu_q(X_p)}{\sum_{p=1}^{m} \mu_q(X_p)}, \qquad (2)$$

$$Supp(A_q \Rightarrow Class\ C_q) = \frac{1}{m} \cdot \sum_{X_p \in Class\ C_q} \mu_q(X_p), \qquad (3)$$

In order to assign a weight to each rule, several heuristic measures proposed in past researches. We have used a heuristic measure proposed in [1] for rule weight specification. This measure calculated by (4).

$$CF_q = conf(A_q \Rightarrow Class\ C_q) - conf_{sum} \qquad (4)$$

Where, $conf_{sum}$ is the sum of confidence of the fuzzy rules having $A_q$ in antecedent part and consequent classes are not $Class_q$. For a $C$-class problem, $conf_{sum}$ can be calculated by (5).

$$conf_{sum} = \sum_{h=1, h \neq Class_q}^{C} conf(A_q \Rightarrow Class\ C_h). \quad (5)$$

The most common reasoning methods are single winner reasoning method and weighted vote reasoning method. In the case of using single winner reasoning method for classifying new patterns (assume the classifier have $R$ rules), the single winner rule $R_w$ is determined by (6) and (7).

$$\mu_w(X_p).CF_w = \max\{\mu_q(X_p).CF_q : q = 1,...,R\}, \quad (6)$$

$$w = \arg\max_q \{\mu_q(X_p).CF_q : q = 1,...,R\} \quad (7)$$

We have generated fuzzy rules with two antecedent conditions and product of confidence and support of rule is used as a certainty grade of the rule. The consequent class of an antecedent combination is specified by finding the class with maximum product of confidence and support. When the consequent class cannot be uniquely determined, the rule is not generated. We have used an evolutionary approach to specify rule weights.

The new pattern $X_p$ is classified as class $C_w$, which is the consequent class of the winner rule $R_w$. If no fuzzy rule covers the $X_p$ and compatible with it or if for $X_p$ multiple fuzzy rules have the same maximum value (product of compatibility grade and certainty grade), but different consequent classes, the classification of $X_p$ is rejected.

## 2. Evaluated T-norms

The intersection of two fuzzy sets $A$ and $B$ is specified by a function $T$: $[0, 1] \times [0, 1] \to [0, 1]$, which aggregates two membership grades by (8).

$$\mu_{A \cap B}(x) = T(\mu_A(x), \mu_B(x)) \quad (8)$$

This class of fuzzy intersection operators are usually referred to as *T*-norm (triangular norm) operators. In mathematics, a *T*-norm is a kind of binary operation used in the framework of probabilistic metric spaces and in multi-valued logic, specifically in fuzzy logic. A *T*-norm generalizes intersection in a lattice and conjunction in logic. The name triangular norm refers to the fact that in the framework of probabilistic metric spaces *T*-norms are used to generalize triangle inequality of ordinary metric spaces. A *T*-norm is a binary operation $T$: $[0, 1] \times [0, 1] \to [0, 1]$ satisfying for all $x, y, z \in [0, 1]$:

1. $T(x, y) = T(y, x)$                        (*T* is commutative)
2. $T(x, T(y, z)) = T(T(x, y), z)$     (*T* is associative)
3. $T(x, 1) = T(1, x) = x$             (1 is an identity)
4. $y \leq z$ implies $T(x, y) \leq T(x, z)$    (*T* is increasing in each variable)

The first requirement indicates that the operator is indifferent to the order of the fuzzy sets to be combined. The second requirement allows us to take the intersection of any number of sets is any order of pair wise groupings. The third requirement shows, 1 are an identity. We have considered 9 different *T*-norms and compared them according to their results in classification accuracy. Table 1 shows specification of these T-norms.

Table 1. Evaluated T-norms.

| Type | T-norm |
|---|---|
| Minimum | $\min(x, y) = \begin{cases} x & \text{if } x \leq y \\ y & \text{if } y < x \end{cases}$ |
| Product | $\prod(x, y) = x \times y$ |
| Yager | $\max\{1 - ((1-x)^\alpha + (1-y)^\alpha)^{1/\alpha}\}$, where $\alpha > 0$ |
| Sugeno-Weber | $\max\left\{\frac{x + y - 1 + \alpha.x.y}{1 + \alpha}\right\}$, where $\alpha \geq -1$ |
| Hamacher | $\frac{x.y}{\alpha + (1-\alpha).(x + y - x.y)}$, where $\alpha \geq 0$ |
| Schweizer-Sklar | $(\max\{x^{-\alpha} + y^{-\alpha} - 1, 0\})^{1/\alpha}$, where $\alpha > 0$ |
| Aczel-Alsina | $e^{-((-\ln x)^\alpha + (-\ln y)^\alpha)^{1/\alpha}}$, where $\alpha > 0$ |
| Dombi | $\frac{1}{1 + ((\frac{1-x}{x})^\alpha + (\frac{1-y}{y})^\alpha)^{1/\alpha}}$, where $\alpha > 0$ |
| Dubois-Prade | $\frac{x.y}{\max(x, y, \alpha)}$, where $\alpha \in [0,1]$ |

## 3. Experiment Results

In this section, we have investigated the effect of 9 different *T*-norm operators (are shown in Table 1) on the accuracy of fuzzy rule-based classification systems. We have examined the classification performance of fuzzy rule-based classification systems designed by using these *T*-norms (in calculating compatibility degree of each pattern with each rule and calculating support, confidence and weight for each rule) through computer simulations. Differences among these *T*-norms are visually demonstrated through experiment results. We have used 12 data sets with numerical attributes from the University of California, Irvine machine learning repository (*UCI*) [2], all of them valid for classification tasks. Table 2 shows specification of these data sets. For each data set the name, number of samples, number of attributes and number of classes are given.

Table 2. Statistics of data sets used in this paper.

| Data set | Number of attributes | Number of samples | Number of classes |
|---|---|---|---|
| Wisconsin (Breast cancer wisconsin) | 10 | 699 | 2 |
| Pima (Pima diabetes) | 8 | 768 | 2 |
| Haberman | 4 | 306 | 2 |
| Heart Statlog | 13 | 270 | 2 |
| Liver (Liver disorders) | 7 | 345 | 2 |
| Labor | 16 | 57 | 2 |
| Wine | 13 | 178 | 3 |
| Thyroid (New Thyroid) | 5 | 215 | 3 |
| Balance (Balance-Scale) | 4 | 625 | 3 |
| Iris | 5 | 150 | 3 |
| Post (Post-Operative Patient) | 8 | 90 | 3 |
| Ecoli | 8 | 336 | 8 |

For experiments, we have employed the ten-fold cross-validation (10-CV) testing method as a validation scheme to perform the experiments and analyze the results. We have run the algorithms five times and the average of accuracies is calculated, for each data set. In ten-fold cross-validation method, each data set is randomly divided into ten disjoint sets of equal size (the size of each set is *m / 10*, where *m* is the total number of patterns in data set). The FRBCS is trained ten times, each time one of ten sets hold out as a test set for evaluating FRBCS and the nine remainder sets are used for training. The classification accuracy is computed in each time and the estimated classifier performance is the average of these 50 classification accuracies.

The experiment results are listed in Table 3. The best results in each row (for each data set) are highlighted by boldface. However, this observation-based evaluation does not reflect whether or not the differences among the methods are significant.

We have used statistical tests to make sure that the difference is significant, that is, big enough that it could not have happened by chance, or in other words, very unlikely to have been caused by chance - the so-called *p*-value of the test [3]. To evaluate the performance of the proposed method, we are used Friedman test [4], which is a non-parametric statistical analysis based on multiple comparison procedures. In order to perform a multiple comparison, it is necessary to check whether all the results obtained by the algorithms present any inequality. Friedman test, ranks the algorithms for each data set separately, the best performing algorithm getting the rank of 1, the second best rank 2, and so on. In case of ties, average ranks are assigned. Under the null-hypothesis, it states that all the algorithms are equivalent, so a rejection of this hypothesis implies the existence of differences among the performance of all the algorithms studied [5]. Friedman's working way of test is described as follows:

Let $r_i^j$ be the rank of the *j-th* of *k* algorithms on the *i-th* of *N* data sets. The Friedman test compares the average ranks of algorithms, $R_j = \frac{1}{N} \sum_i r_i^j$. Under the null-hypothesis, which states that all the algorithms are equivalent and so their ranks $R_j$ should be equal, the Friedman statistic is distributed according to $\chi_F^2$ with $k - 1$ degrees of freedom and is as follows [6]:

$$\chi_F^2 = \frac{12N}{k(k+1)} \cdot [\sum_j R_j^2 - \frac{k(k+1)^2}{4}]. \tag{9}$$

Average ranks obtained by each method in the Friedman test are shown in Table 4. In this table, the value of Friedman statistic (distributed according to chi-square with 8 degrees of freedom) is 25.283333 and *p*-value computed by this test is 0.00139164000876435. These rank values will be useful to calculate the *p*-values and to detect significant differences between the methods. Evidently, the ranks assigned to Aczel-Alsina, Dubois-Prade and Dombi are less than other *T*-norm operator's ranks. Hence, Aczel-Alsina, Dubois-Prade and Dombi are the best performing *T*-norm operators. The results, which are analyzed by statistical techniques, correspond to average accuracies in test data.

Table 3. Comparing the classification accuracy (10-CV test method).

| T-norm / Data set | Minimum | Product | Yager | Sugeno-Weber | Hamacher | Schweizer-Sklar | Aczel-Alsina | Dombi | Dubois-Prade |
|---|---|---|---|---|---|---|---|---|---|
| **Pima** | 69.40 | 69.40 | 71.00 | 65.11 | 52.42 | 70.69 | **72.87** | 72.18 | 70.37 |
| **Haberman** | 73.11 | 72.76 | 72.17 | **73.56** | 73.00 | 73.03 | **73.56** | 73.10 | 73.40 |
| Liver | 57.97 | 58.20 | **59.16** | 57.97 | 51.00 | 57.49 | 58.54 | 57.76 | 58.17 |
| Labor | 84.08 | 85.33 | 80.46 | 65.36 | 77.51 | **90.50** | 84.94 | 81.63 | 77.51 |
| Thyroid | 88.91 | 88.72 | 91.93 | 69.86 | 88.80 | 91.63 | **92.71** | 91.68 | 91.41 |
| Balance | 89.37 | 89.34 | 88.75 | 87.53 | 89.87 | **90.08** | 89.86 | 89.84 | 89.94 |
| Iris | 96.00 | 95.60 | 95.46 | 95.33 | 96.00 | **96.80** | 95.86 | 95.33 | 96.00 |
| Post | 72.38 | **73.51** | 72.62 | 71.11 | 73.17 | 70.22 | 73.13 | 73.17 | 73.17 |
| Wisconsin | 94.91 | 95.07 | 95.26 | 80.30 | 94.93 | **96.17** | 95.77 | 95.56 | 96.13 |
| Heart | 79.93 | 80.46 | 79.61 | 73.62 | **81.10** | 78.22 | 79.12 | 80.34 | **81.10** |
| Wine | 92.90 | 92.67 | 93.83 | 78.93 | 92.97 | 94.88 | 95.36 | **96.09** | 93.09 |
| Ecoli | 74.51 | 73.31 | 75.52 | 60.01 | 66.74 | 73.11 | 75.54 | 75.86 | **76.64** |

Table 4. Average rankings of algorithms by Friedman procedure.

| Algorithm | Ranking |
|---|---|
| **Minimum** | 5.625 |
| **Product** | 6.4375 |
| **Yager** | 4.375 |
| **Sugeno-Weber** | 7.4375 |
| **Hamacher** | 7 |
| **Schweizer-Sklar** | 4.25 |
| **Aczel-Alsina** | 2.3125 |
| **Dombi** | 4.0625 |
| **Dubois-Prade** | 3.5 |

## 4. CONCLUSIONS

This paper presents a comparative study which examines a number of *T*-norms and their effect on the accuracy of fuzzy rule-based classification systems. We have used confidence and support for selecting and weighting the fuzzy rules. Usually, for calculating these merits the product has been used as a *T*-norm. But, for calculating these merits we used 9 different *T*-norm operators. Simulation results on 12 well-known data sets showed that employing Aczel-Alsina operator can improve the performance of the classification. The second rank stands for Dubois-Prade followed by the third rank stands for Dombi.